%% file: root.tex
\PassOptionsToPackage{unicode}{hyperref}
\PassOptionsToPackage{hyphens}{url}
\PassOptionsToPackage{dvipsnames,svgnames,x11names}{xcolor}
\documentclass[
  letterpaper,
  10pt,
  conference]{IEEEtran}
\usepackage{xcolor}
\usepackage{amsmath,amssymb}
\usepackage{iftex}
\ifPDFTeX
  \usepackage[T1]{fontenc}
  \usepackage[utf8]{inputenc}
  \usepackage{textcomp} 
\else 
  \usepackage{unicode-math} 
  \defaultfontfeatures{Scale=MatchLowercase}
  \defaultfontfeatures[\rmfamily]{Ligatures=TeX,Scale=1}
\fi
\usepackage{lmodern}
\IfFileExists{upquote.sty}{\usepackage{upquote}}{}
\IfFileExists{microtype.sty}{
  \usepackage[]{microtype}
  \UseMicrotypeSet[protrusion]{basicmath} 
}{}
\makeatletter
\@ifundefined{KOMAClassName}{
  \IfFileExists{parskip.sty}{%
    \usepackage{parskip}
  }{
    \setlength{\parindent}{0pt}
    \setlength{\parskip}{6pt plus 2pt minus 1pt}}
}{
  \KOMAoptions{parskip=half}}
\makeatother
\makeatletter
\ifx\paragraph\undefined\else
  \let\oldparagraph\paragraph
  \renewcommand{\paragraph}{
    \@ifstar
      \xxxParagraphStar
      \xxxParagraphNoStar
  }
  \newcommand{\xxxParagraphStar}[1]{\oldparagraph*{#1}\mbox{}}
  \newcommand{\xxxParagraphNoStar}[1]{\oldparagraph{#1}\mbox{}}
\fi
\ifx\subparagraph\undefined\else
  \let\oldsubparagraph\subparagraph
  \renewcommand{\subparagraph}{
    \@ifstar
      \xxxSubParagraphStar
      \xxxSubParagraphNoStar
  }
  \newcommand{\xxxSubParagraphStar}[1]{\oldsubparagraph*{#1}\mbox{}}
  \newcommand{\xxxSubParagraphNoStar}[1]{\oldsubparagraph{#1}\mbox{}}
\fi
\makeatother

\usepackage{longtable,booktabs,array}
\usepackage{calc} 
\usepackage{etoolbox}
\makeatletter
\patchcmd\longtable{\par}{\if@noskipsec\mbox{}\fi\par}{}{}
\makeatother
\IfFileExists{footnotehyper.sty}{\usepackage{footnotehyper}}{\usepackage{footnote}}
\makesavenoteenv{longtable}
\usepackage{graphicx}
\makeatletter
\newsavebox\pandoc@box
\newcommand*\pandocbounded[1]{
  \sbox\pandoc@box{#1}%
  \Gscale@div\@tempa{\textheight}{\dimexpr\ht\pandoc@box+\dp\pandoc@box\relax}%
  \Gscale@div\@tempb{\linewidth}{\wd\pandoc@box}%
  \ifdim\@tempb\p@<\@tempa\p@\let\@tempa\@tempb\fi
  \ifdim\@tempa\p@<\p@\scalebox{\@tempa}{\usebox\pandoc@box}%
  \else\usebox{\pandoc@box}%
  \fi%
}
\def\fps@figure{htbp}
\makeatother

\providecommand{\tightlist}{%
  \setlength{\itemsep}{0pt}\setlength{\parskip}{0pt}}

\usepackage[numbers]{natbib}
\IEEEoverridecommandlockouts
\usepackage[letterpaper, top=1in, bottom=0.75in, left=0.75in, right=0.75in]{geometry}
\usepackage{etoolbox}
\newcounter{marginfixed}
\AddToHook{shipout/before}{%
  \ifnum\value{marginfixed}=0
    \stepcounter{marginfixed}%
  \else
    \ifnum\value{marginfixed}=1
      \stepcounter{marginfixed}%
      \global\advance\topmargin by -0.25in
      \global\advance\textheight by 0.25in
    \fi
  \fi
}
\usepackage{tikz}
\usetikzlibrary{positioning,shapes.geometric,calc,arrows.meta}
\usepackage{booktabs}
\AtBeginDocument{%
  \setlength{\parskip}{0pt}%
  \setlength{\parindent}{1em}%
}
\makeatletter
\@ifpackageloaded{caption}{}{\usepackage{caption}}
\AtBeginDocument{%
\ifdefined\contentsname
  \renewcommand*\contentsname{Table of contents}
\else
  \newcommand\contentsname{Table of contents}
\fi
\ifdefined\listfigurename
  \renewcommand*\listfigurename{List of Figures}
\else
  \newcommand\listfigurename{List of Figures}
\fi
\ifdefined\listtablename
  \renewcommand*\listtablename{List of Tables}
\else
  \newcommand\listtablename{List of Tables}
\fi
\ifdefined\figurename
  \renewcommand*\figurename{Figure}
\else
  \newcommand\figurename{Figure}
\fi
\ifdefined\tablename
  \renewcommand*\tablename{Table}
\else
  \newcommand\tablename{Table}
\fi
}
\@ifpackageloaded{float}{}{\usepackage{float}}
\floatstyle{ruled}
\@ifundefined{c@chapter}{\newfloat{codelisting}{h}{lop}}{\newfloat{codelisting}{h}{lop}[chapter]}
\floatname{codelisting}{Listing}

\makeatother
\makeatletter
\@ifpackageloaded{caption}{}{\usepackage{caption}}
\@ifpackageloaded{subcaption}{}{\usepackage{subcaption}}
\makeatother
\usepackage{bookmark}
\IfFileExists{xurl.sty}{\usepackage{xurl}}{} 
\hypersetup{
  colorlinks=true,
  linkcolor={blue},
  filecolor={Maroon},
  citecolor={Blue},
  urlcolor={Blue},
  pdfcreator={LaTeX via pandoc}}

\author{}
\date{}
\begin{document}

\title{\vspace*{-9mm} \LARGE \bfseries\itshape How Should I Pick a Foundation Model for My Robot?\\ \Large \upshape In Favor of a Community Evaluation Framework for Social Robots\vspace*{-3mm}}
\author{Eric Nichols$^{1}$, Alva Markelius$^{2}$, and Hatice Gunes$^{2}$%
\thanks{Accepted at \href{https://sites.google.com/cam.ac.uk/forma/}{FoRMA:
Foundation Models in the RO-MAN Age --- Responsible Development for Social
Robotics}, a workshop at IEEE RO-MAN 2026, Kitakyushu, Japan.}%
\thanks{$^{1}$Honda Research Institute Japan, Wako, Japan.
{\tt\small e.nichols@jp.honda-ri.com}}%
\thanks{$^{2}$Affective Intelligence and Robotics Lab, University of Cambridge, UK.
{\tt\small \{ajkm4,hg410\}@cam.ac.uk}}%
}
\maketitle
\thispagestyle{empty}
\pagestyle{empty}

\section*{Abstract}\label{abstract}
\addcontentsline{toc}{section}{Abstract}

Researchers who seek to build social robot applications on foundation
models are faced with a difficult question: \emph{how should we pick a
model?} Public leaderboards offer little guidance: the demands of
real-time, embodied social interaction lie largely outside their focus.
And direct evaluation is impractical at scale: each embodied study
requires scarce participant, robot, and experimenter time. In this
paper, we identify five evaluation dimensions for foundation models in
social robots: (i) \emph{conversational competence}, (ii) \emph{user
safety}, (iii) \emph{embodied character}, (iv) \emph{target scene
effectiveness}, and (v) \emph{audience appropriateness}. To make model
selection cheaper and better informed, we propose a three-tiered
\emph{evaluation funnel paradigm} that first filters with general
metrics, then extends to simulated interactions, and terminates in more
expensive, robot-specific evaluation. We map all five dimensions across
all three tiers, chart where applicable evaluation methods exist and are
missing, and close with a call to action: \emph{let's build the
evaluation framework together as a community}.

\section{The Problem}\label{sec-problem}

Foundation models power a growing variety of social robot applications
\citep{zhang2023llmhri}. Recent examples include open-domain companion
dialogue with older adults \citep{irfan2025reality}, robot-led wellbeing
assessment of children \citep{abbasi2025vlm}, and expressive
conversational behavior generation \citep{wang2024maru}. However,
choosing which model to build on is hard: candidate models are many, and
the field moves quickly. Integration binds prompts, pipelines, and
guardrails to the chosen model, making the choice costly to reverse. As
a result, the choice is often made early, on thin evidence, because
stronger evidence is costly to collect.

The obvious source of evidence is public leaderboards and benchmarks.
However, prominent leaderboards such as LM Arena
\citep{chiang2024chatbotarena} and LiveBench \citep{white2025livebench}
focus on skills that barely overlap with the needs of a social robot.
The benchmark suites that headline model releases reward specialized
reasoning: programming
(\href{https://arxiv.org/abs/2107.03374}{HumanEval}), mathematics
(\href{https://arxiv.org/abs/2110.14168}{GSM8K}), and exam-style problem
solving (\href{https://arxiv.org/abs/2501.14249}{Humanity's Last Exam}).
These skills matter more to software agents than to embodied social
agents.

Evaluation conditions pose a second problem. Public scores are measured
at full precision, with no latency budget. Human turn transitions
cluster around 200 ms \citep{sacks1974, stivers2009}, leaving a robot at
most a second to respond before disconnect is noticed. This budget
leaves no time for the heavy chain-of-thought processing that many
benchmark scores depend on.

Interactive benchmarks come closer to what we need. SOTOPIA
\citep{zhou2023sotopia} stages multi-turn social scenarios between LLM
agents and scores them with an LLM judge; its scenario-and-judge
machinery is useful. However, SOTOPIA is text-only: no body, no
real-time pressure, no robot persona. RoboArena
\citep{liang2025roboarena} compares real robot policies through
distributed, crowd-judged head-to-head trials, and its
community-distributed evaluation model is what embodied evaluation
needs, but it measures manipulation success rather than social
interaction. Between these two frameworks lies the uncovered ground of
embodied social interaction. The relevant pieces exist, scattered;
nothing assembles them for social robots. Our scope is accordingly the
model as social interlocutor (conversation, expression, and persona),
not foundation models for manipulation or navigation, which pose
different evaluation problems.

Indeed, existing benchmarks remain largely non-interactive and
non-embodied, leaving the socio-technical, affective, and social
capabilities that matter for social robots, where the meaning of success
is co-constructed through context, embodiment, and shared norms,
insufficiently evaluated \citep{markelius2026desiderata}. So, do we even
know what a social robot needs us to evaluate, and how? From the
deployment scenes above and the demands they place on a model, we
distill five dimensions of evaluation for social robots:

\begin{enumerate}
\def\labelenumi{\arabic{enumi}.}
\tightlist
\item
  \textbf{Conversational competence.} The model must sustain coherent,
  engaging, multi-turn conversation, and it must do so in real time.
\item
  \textbf{User safety.} The model must do no harm: it must refuse
  dangerous requests, avoid harmful content, and not foster dependency.
\item
  \textbf{Embodied character.} The model must speak and act as the
  target robot, staying in persona and within the platform's expressive
  repertoire.
\item
  \textbf{Target scene effectiveness.} The model must advance the goals
  of its deployment scene, whether tutoring a classroom, coaching
  wellbeing, or explaining a diagnosis.
\item
  \textbf{Audience appropriateness.} The model's language, topics, and
  interaction style must fit the population it serves: children, older
  adults, patients.
\end{enumerate}

These five dimensions can be expanded into finer, separately measurable
capabilities. For example, Markelius et al.'s desiderata
\citep{markelius2026desiderata} are organized across situated,
relational, and knowledge layers, and offer one such expansion: they
decompose user safety into proactive harm avoidance, emotional
sensitivity, and non-sycophancy, and audience appropriateness into
cultural localization and epistemic heterogeneousness. We return to
these expansions in Section~\ref{sec-together}.

No existing benchmark measures all five dimensions. Furthermore, the
applicable evaluation methods differ sharply in \emph{cost}: static
tests are cheap, while embodied studies are precious. Ordered from cheap
to expensive, each stage prunes the candidate pool, maximizing
information about the final pick while minimizing its cost. We formalize
this approach as a three-tiered \emph{evaluation funnel} in
Section~\ref{sec-funnel}.

Our contributions are as follows: (i) five evaluation dimensions for
foundation models in social robots; (ii) an evaluation funnel
architecture with a coverage map charting where applicable methods exist
and are missing; and (iii) a call to build the evaluation framework
together.

\input{tbl_map.tex}

\section{An Evaluation Funnel}\label{sec-funnel}

\begin{figure}[!t]

\centering{

\includegraphics[width=1\linewidth,height=\textheight,keepaspectratio]{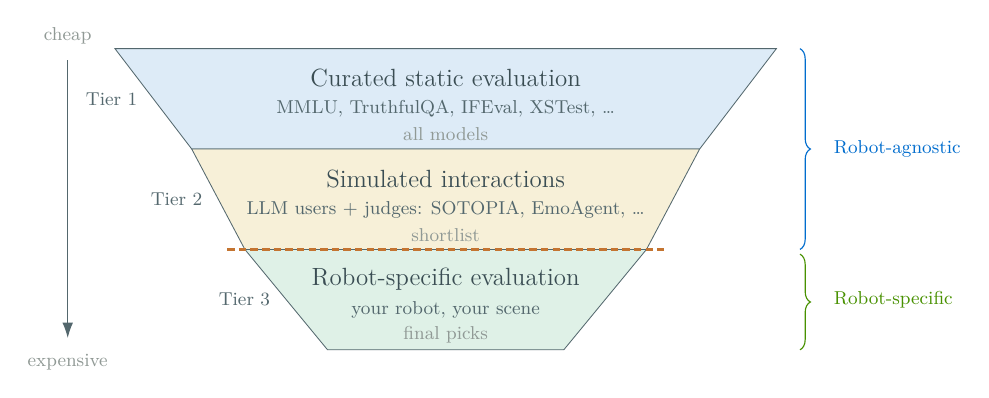}

}

\caption{\label{fig-funnel}The evaluation funnel. Candidate models
narrow through three tiers as evaluation cost and specificity grow.}

\end{figure}%

Figure~\ref{fig-funnel} shows the funnel, where candidate models enter
wide, narrow tier by tier, and exit as final picks:

\begin{enumerate}
\def\labelenumi{\arabic{enumi}.}
\tightlist
\item
  \textbf{Tier 1: \emph{curated static evaluation}} filters with
  benchmarks selected to inform competence, safety, and character.
\item
  \textbf{Tier 2: \emph{simulated interactions}} uses LLM-based users
  and judges to test these dimensions further and to extend coverage to
  target scenes and audiences.
\item
  \textbf{Tier 3: \emph{robot-specific evaluation}} begins with
  harness-based simulation before deployment, reducing live testing.
\end{enumerate}

Tiers 1 and 2 are robot-agnostic and can be shared as a living community
leaderboard, while Tier 3 harnesses are platform-specific. None of this
is hypothetical: Tier 1 can be assembled today, and Tier 2 can build on
existing multi-turn benchmarks.

Each candidate yields per-dimension performance scores together with
deployment-cost measurements: VRAM, latency, and parameter count. In
addition to a traditional leaderboard ranking, the evaluation funnel can
therefore include a \emph{Pareto frontier}
\citep{liang2022helm, recasens2024}: the strongest candidate at each
cost point. The frontier is computed per dimension over the performance
versus deployment-cost trade-off, under a real-time latency cutoff. Each
researcher then picks the frontier point that fits their deployment
envelope, rather than a single global winner.

\subsection{Tier 1: Curated Static
Evaluation}\label{tier-1-curated-static-evaluation}

Tier 1 curates existing benchmarks rather than building new ones: the
field already offers hundreds, so the work is selection. Curation keeps
suites that inform our dimensions, including general knowledge (MMLU
\citep{hendrycks2021mmlu}), commonsense and social reasoning
(CommonsenseQA \citep{talmor2019commonsenseqa}, HellaSwag
\citep{zellers2019hellaswag}, Social IQa \citep{sap2019socialiqa}),
truthfulness as a safety property (TruthfulQA
\citep{lin2022truthfulqa}), instruction following (IFEval
\citep{zhou2023ifeval}), and refusal calibration (XSTest
\citep{rottger2024xstest}, OR-Bench \citep{cui2024orbench}). It drops
suites that reward the slow chain-of-thought a real-time budget cannot
afford (\href{https://arxiv.org/abs/2110.14168}{GSM8K}, BBH
\citep{suzgun2022bbh}).

Everything runs under deployment conditions: quantized weights and
capped response time. Mature open infrastructure such as the lm-eval
harness \citep{lmevalharness} and vLLM serving \citep{kwon2023vllm}
makes this curation largely a configuration task rather than new code.
Furthermore, Tier 1 is cheap enough to rerun on every model release, so
the leaderboard stays current as the field moves.

\subsection{Tier 2: Simulated
Interactions}\label{tier-2-simulated-interactions}

Static suites cannot test multi-turn behavior in context: persona hold,
scene goals, audience fit. Tier 2 therefore stages simulated
conversations conditioned on character, scene, and audience. LLM-based
users role-play the target audience
\citep{zhou2023sotopia, emoagent2025}, and LLM-based judges score the
resulting transcripts against per-dimension rubrics
\citep{gu2024llmjudgesurvey}. Judges can reach human-level agreement on
conversational quality \citep{zheng2023judge} and on safety in sensitive
domains \citep{bentley2026veramh}. However, judges favor their own
generations \citep{panickssery2024self}, and simulation without
information asymmetry inflates social competence
\citep{zhou2024misleading}. Judge scores are therefore priors, not
verdicts; optimizing against a fixed judge drifts it from human judgment
\citep{wang2024sotopiapi}. In exchange, Tier 2 could buy thousands of
in-context conversations per candidate before human evaluation, paying
in tokens rather than participant time.

\subsection{Tier 3: Robot-Specific
Evaluation}\label{tier-3-robot-specific-evaluation}

Tier 3 answers what only embodiment can: expressive behavior on
hardware, the real scene, the real population. Even here, evaluation
need not begin live: a platform harness can replay Tier 2 scenarios
through the robot's perception and behavior stack, catching integration
failures before any participant is recruited. The funnel's job is to
keep this tier small, so that user studies spend their scarce
participant and experimenter time on the research question rather than
on model comparison.

Table~\ref{tbl-map} maps the five dimensions against the three tiers.
The coverage is uneven: conversational competence is well-served
cheaply, while embodied character, scene, and audience are not. The
framework's job is to push every dimension as far up the funnel as
honesty allows, and the map's empty cells show where to begin.

\section{A Representative Case Study}\label{sec-case}

Consider Haru, an expressive tabletop social robot
\citep{gomez2018haru}, facilitating English-communication practice with
small groups of Japanese high-school students. Haru acts as a
peer-adjacent facilitator: it encourages participation and turn-taking
and draws reticent students into the conversation. The scene
instantiates all five dimensions at once: real-time multi-party
conversation (\emph{conversational competence}), safety guardrails
appropriate to minors (\emph{user safety}), Haru's established playful
persona (\emph{embodied character}), pro-social goals (\emph{target
scene effectiveness}), and adolescent language learners (\emph{audience
appropriateness}).

Recent releases of popular open-weight model families, in sizes that run
on consumer hardware, yield a candidate pool of 30--40 models. Tier 1
computes the Pareto frontier over performance and deployment cost,
locating the strongest candidates at this classroom's operating point
(e.g., 24 GB of VRAM and sub-2 s replies). Its curated gates supply
per-dimension evidence from existing benchmarks: Social IQa probes
whether a model reads social situations, and XSTest and OR-Bench check
that refusals are strict enough for minors without over-refusing benign
classroom talk.

Tier 2 then runs the shortlisted models in an agentic simulation of the
classroom: each candidate actually converses with LLM users role-playing
three to four English learners, one of whom stays silent, while judges
score persona hold, goal inference and adherence to the scenario, and
whether the model yields turns and draws the quiet student in. Tier 3
reaches what only the platform can answer: the final picks drive Haru's
expressive behavior (animated routines, gaze, and tone of voice) through
its behavior stack under the same latency budget. A harness simulating
real-time multimodal input could extend evaluation to full physical
behavior, an open question we return to in Section~\ref{sec-together}.

However, a per-candidate user study remains infeasible in a
curriculum-embedded, longitudinal deployment: school terms move slower
than model releases. The funnel instead delivers one or two vetted
models, and that set is fixed before any student meets the robot. No
model changes mid-study, so the pilot's scarce supervised sessions
measure what matters: participation and turn-taking dynamics, the
questions the study exists to answer.

\section{Building It Together}\label{sec-together}

Choosing a foundation model for a social robot remains hard: many
candidates, multiple dimensions, and evaluation that grows most
expensive exactly where it matters most. Today, everyone faces this
problem alone. However, nothing in Tiers 1 and 2 is robot-specific.
Built once and shared as a living community leaderboard, they would give
everyone the same starting evidence. A shared platform buys
cross-project and cross-robot comparability, and community curation
resists benchmark decay better than proliferation does
\citep{reuel2024betterbench}. The upper tiers also deliver information
early: per-dimension evidence arrives before any hardware is touched.

We do not pretend the last mile is cheap: harnesses and user studies are
robot-specific, and no shared leaderboard can absorb them. However,
building Tier 3 on the upper tiers automates the synthetic part: Tier 2
scenarios replay through the platform harness, stress-testing candidates
before any deployment. Failures then surface at the cheapest tier that
can catch them, and every check that runs upstream spares a vulnerable
user downstream. Per-robot harnesses start from the leaderboard's
shortlists, and the community can share harnesses for popular platforms.
Every shared harness lowers the cost of entry for the next robot.

Operationalizing Tier 3 leaves open questions we put to the community.
First, we have largely treated candidates as language models, but
embodied perception increasingly runs through vision-language models
\citep{abbasi2025vlm}: how to fold multimodal interaction into the
funnel, and how far up it can be pushed before embodiment becomes
unavoidable, is unresolved. Second, Tier 3 combines two instruments of
unequal cost, i.e., harness replay and live user studies, and the
boundary between them is unsettled: knowing which social phenomena a
harness can validly stand in for, and which require a user, would tell
us when harness evidence is enough and when it must defer to a user
study. Third, robot-specificity is a matter of degree. The harness is
platform-bound by construction, but its scenario logic, perception
interfaces, and scoring rubrics need not be; finding the abstractions
that let a harness built for one platform inform another would push the
funnel's shared, robot-agnostic reach further down than we have assumed.

Markelius et al.'s evaluation desiderata \citep{markelius2026desiderata}
approach the same problem bottom-up, from the needs of vulnerable
populations. The two trajectories meet in the middle: Tiers 1 and 2 hand
such a project a defensible model shortlist, and its datasets are what
the lower funnel is made of. Our five dimensions are the
practitioner-facing projection of what the desiderata demand.

So how should you pick a foundation model for your robot? Pick your
point on the Pareto frontier. You get a Pareto frontier by building the
framework together.

\bibliography{references_short.bib}

\end{document}

%% file: tbl_map.tex
\begin{table*}[!t]
\caption{\label{tbl-map}Coverage map: public benchmarks applicable to each dimension and tier.
Entries are exemplars, not an exhaustive list; empty cells are the community's open problems.}
\centering
\footnotesize
\setlength{\tabcolsep}{3pt}
\begin{tabular}{@{}llll@{}}
\toprule
Dimension & T1: curated static evaluation & T2: simulated interactions & T3: robot-specific evaluation \\
\midrule
Competence &
  \href{https://arxiv.org/abs/2009.03300}{MMLU},
  \href{https://aclanthology.org/N19-1421/}{CommonsenseQA},
  \href{https://aclanthology.org/P19-1472/}{HellaSwag},
  \href{https://aclanthology.org/D19-1454/}{Social IQa},
  \href{https://arxiv.org/abs/2311.07911}{IFEval} &
  \href{https://arxiv.org/abs/2310.11667}{SOTOPIA} &
  \href{https://arxiv.org/abs/2504.13898}{SHREC} \\
Safety &
  \href{https://aclanthology.org/2022.acl-long.229/}{TruthfulQA},
  \href{https://aclanthology.org/2024.naacl-long.301/}{XSTest},
  \href{https://proceedings.mlr.press/v267/cui25a.html}{OR-Bench} &
  \href{https://arxiv.org/abs/2504.09689}{EmoAgent},
  \href{https://arxiv.org/abs/2602.05088}{VERA-MH} &
  \href{https://arxiv.org/abs/2606.29937}{REPAIR-Bench} \\
Character &
  \href{https://arxiv.org/abs/2310.00746}{RoleLLM},
  \href{https://arxiv.org/abs/2406.14703}{TRAIT} &
  \href{https://arxiv.org/abs/2407.18416}{PersonaGym} &
  --- \\
Scene &
  \href{https://arxiv.org/abs/1810.00278}{MultiWOZ} &
  \href{https://arxiv.org/abs/2310.11667}{SOTOPIA} &
  --- \\
Audience &
  \href{https://arxiv.org/abs/2503.10242}{MinorBench} &
  \href{https://arxiv.org/abs/2504.09689}{EmoAgent} &
  --- \\
\bottomrule
\end{tabular}
\end{table*}